\pdfoutput=1

\documentclass[11pt]{article}

\usepackage[]{ACL2023}

\usepackage{times}
\usepackage{latexsym}
\usepackage{amssymb}
\usepackage{amsmath}
\usepackage{graphicx}
\graphicspath{{images/}}

\usepackage{amsthm,amsmath,amssymb}
\usepackage{bbm}

\usepackage{mathrsfs}
\usepackage{booktabs}
\usepackage{multirow}

\usepackage[T1]{fontenc}

\usepackage[utf8]{inputenc}

\usepackage{microtype}

\usepackage{inconsolata}

\newcommand\blfootnote[1]{%
  \begingroup
  \renewcommand\thefootnote{}\footnote{#1}%
  \addtocounter{footnote}{-1}%
  \endgroup
}

\title{
    Non-parametric, Nearest-neighbor-assisted Fine-tuning for\\ Neural Machine Translation
}

\author{Jiayi Wang$^\ast$\textsuperscript{1}, Ke Wang$^\ast$\textsuperscript{2}, Yuqi Zhang\textsuperscript{2}, Yu Zhao\textsuperscript{2}, Pontus Stenetorp\textsuperscript{1} \\
\textsuperscript{1}University College London\\
\textsuperscript{2}Alibaba DAMO Academy \\
\texttt{ucabj45@ucl.ac.uk,\{wk258730,chenwei.zyq\}@alibaba-inc.com,}\\
\texttt{kongyu@taobao.com,p.stenetorp@cs.ucl.ac.uk} \\
}

\begin{document}
\maketitle
\blfootnote{$^\ast$ Equal Contribution.}

\begin{abstract}
Non-parametric, $k$-nearest-neighbor algorithms have recently made inroads to assist generative models such as language models and machine translation decoders.
We explore whether such non-parametric models can improve machine translation models at the fine-tuning stage by incorporating statistics from the $k$NN predictions to inform the gradient updates for a baseline translation model.
There are multiple methods which could be used to incorporate $k$NN statistics and we investigate gradient scaling by a gating mechanism, the $k$NN's ground truth probability, and reinforcement learning.
For four standard in-domain machine translation datasets, compared with classic fine-tuning, we report consistent improvements of all of the three methods by as much as $1.45$ BLEU and $1.28$ BLEU for German-English and English-German translations respectively. 
Through qualitative analysis, we found particular improvements when it comes to translating grammatical relations or function words, which results in increased fluency of our model.
%

%
%
%
%
%
%
\end{abstract}

\section{Introduction}
%


Non-parametric nearest neighbor models have been seen recent success for generative natural language processing tasks such as language modeling~\citep{khandelwal2020generalization} and machine translation~\citep{Khandelwal2020NearestNM}.
Not only because explicitly memorizing the training data helps generalization, generative natural language models can scale to larger text collections without the added cost of training.
\citet{khandelwal2020generalization} introduced $k$-nearest-neighbor machine translation~($k$NN-MT): a simple non-parametric method for machine translation~(MT) via nearest-neighbor retrievals was proposed and has been verified its effectiveness -- improving BLEU scores by roughly 3  for translating from
English into German and Chinese.
%

To easily adapt to multi domains, during inference, $k$NN-MT interpolates the softmax distribution for the target token from the neural machine translation~(NMT) model with the distribution of the retrieved set generated by the $k$-nearest-neighbor~($k$NN) search on a datastore of cached examples.
The datastore is constructed from key-value pairs of parallel training data, where the key is the latent contextual representation of the target prefix tokens obtained via the NMT's stochastic forward-pass computing, and the value is the corresponding ground-truth target token.
%
%

During preliminary investigations, we observed that the $k$NN search is able to memorize content words with lexical meanings of in-domain contexts. However, when it comes to translate grammatical relations, such as function word translations, querying the datastore for $k$ nearest neighbors is insufficient, which has a negative impact on the fluency of the final translation result.

\begin{table*}[h]
\normalsize
  \centering
  \resizebox{\hsize}{!}
{
  \begin{tabular}{l| c c c c | c | c c c c | c }
  \toprule[1pt]
   &  \multicolumn{5}{c|}{\textbf{De-En}} & \multicolumn{5}{c}{\textbf{En-De}}
    \\
    \midrule[0.75pt]
   \textbf{Model} & \textbf{IT}& \textbf{Medical} & \textbf{Law} & \textbf{Koran} &\textbf{Avg.} & \textbf{IT}& \textbf{Medical} & \textbf{Law} & \textbf{Koran} &\textbf{Avg.}
  \\
  
 \midrule[0.75pt]
 {Base MT}  & 
 38.35 & 40.14 & 45.63 & 16.29 & 35.10 & 29.74 & 35.56 & 40.85 & 13.97 & 30.03
 \\
  {$k$NN-MT}  & 
 46.12 & 54.41 & 61.70 &	21.14 & 45.84 & 36.44 & 49.74 & 55.73 & 25.87 & 41.95
 \\
 {Fine-tuned MT}  &
 47.14 & 57.19 & 61.28 & 22.98 & 47.15 & 39.70 & 52.50 & 57.16 & 32.45 & 45.45
 \\
 {$k$NN-FT-MT}  &
 49.33 & 57.46 & 63.63 & 22.95 & 48.34 & 40.68 & 53.28 & 58.91 & 32.61 & 46.37
 \\
   \bottomrule[1pt]
  \end{tabular}
}
  \caption{Performances of Base NMT and the fine-tuned NMT with and without the integration with $k$NN search during inference in German-English (De-En) and English-German (En-De) multi-domain translations respectively. Results are reported with the metric SacreBLEU \citep{post-2018-call}. $k$NN-MT represents the Base NMT with integration with $k$NN search during inference, while $k$NN-FT-MT represents the fine-tuned NMT with integration with $k$NN search during inference. }
  \label{tab:ft_knn_mt}
\end{table*}

Moreover, although $k$NN-MT has the advantage that it does not require additional fine-tuning, our experiments show that $k$NN-MT cannot outperform or even achieve comparable performance to classic fine-tuning~\cite{mou-etal-2016-transferable} when  in-domain data is accessible.
This is due to the machine translation model not having been optimized on the in-domain data, and thus it limits the translation model's capability to utilize the $k$NN search.
A simple way to observe this fact is to apply the $k$NN-MT algorithm on a fine-tuned translation model, but not a baseline translation model trained on out-of-domain data.
As shown in Table~\ref{tab:ft_knn_mt} and Table~\ref{tab:prob} in Appendix \ref{sec:appendix}, the performance of the algorithm can be largely improved when the $k$NN datastore is constructed with fine-tuned contextual representations and their corresponding keys.
Therefore, fine-tuning is still necessary and it benefits the non-parametric $k$NN search algorithm.
%



Given that there are both advantages and disadvantages stemming from the $k$NN-MT algorithm, we proceed to maximize the use of the results from the $k$NN search to enhance the performance of translation models.
We hypothesize that the fine-tuning procedure of a neural translation model can be improved with the assistance of statistics from the $k$NN predictions.
Furthermore, we also explore gradient scaling for the original neural translation model with (1) a gate mechanism applied on the distribution of $k$NN predictions, (2) the $k$NN ground truth probability and (3) reinforcement learning based on the statistics of $k$NN predictions. 
%


Based on these observations, we propose \textit{trainable}-$k$NN-MT to alleviate the problems of the the original $k$NN-MT~\citep{Khandelwal2020NearestNM}.
Our \textit{trainable}-$k$NN-MT is able to learn translations conditioned on the retrieved k-nearest-neighbors.
The statistics of the retrieved set are incorporated into model fine-tuning via three ways aforementioned to dynamically scale up the gradient for back-propagation.
In addition, the $k$NN datastore for retrieving is jointly updated with model fine-tuning so that the $k$NN search can secure more accurate $k$ nearest neighbors.
%



There are two main contributions in this paper:
(1) The \textit{trainable}-$k$NN-MT generates better objective contextual representations of relevant examples, which improves the retrieved sets of top-k nearest neighbors.
(2) The \textit{trainable}-$k$NN-MT significantly outperforms both of the original $k$NN-MT algorithm and the classic fine-tuning, making it a novel fine-tuning method for neural machine translation. In addition, the fluency of the translation from \textit{trainable}-$k$NN-MT is qualitatively improved, while staying more faithful to the original language.
%
%


\begin{figure*}
\centering
\includegraphics[width=0.9\textwidth]{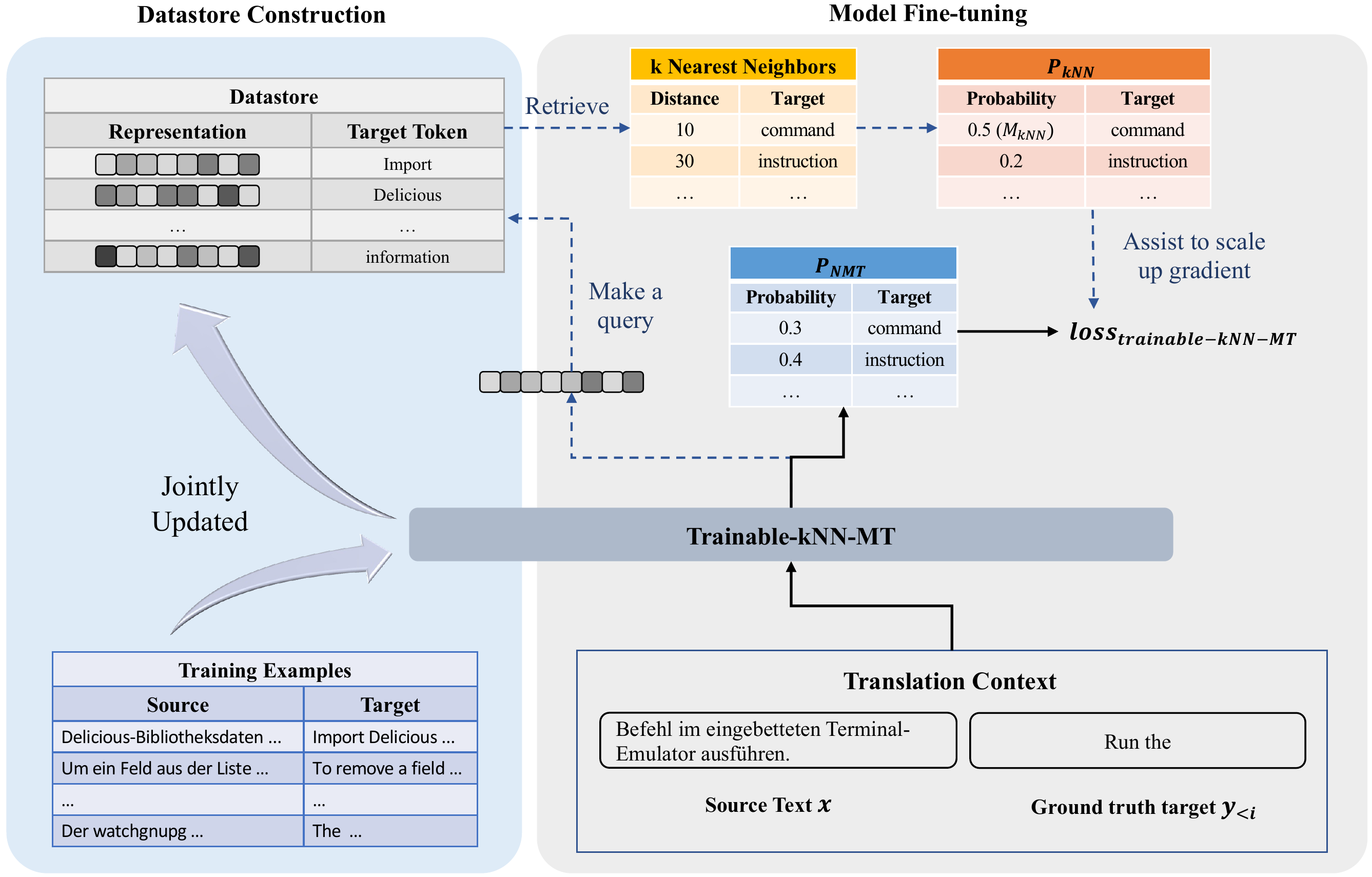}
\caption{The schematic representation of the \textit{trainable}-$k$NN-MT. Arrows with broken lines illustrate the workflow of learning translations with the assistance of statistics from $k$NN predictions. 
}
\label{framework}
\end{figure*}

\section{Methodology}
\label{Method}
In this section, we will introduce the \textit{trainable}-$k$NN-MT model, which is able to (1) generate better objective contextual representations, (2) improve the performances of both of the vanilla fine-tuning and $k$NN-MT algorithm via gradient scaling with the assistance of statistics from $k$NN predictions. 

\subsection{Preliminary Method}
Formerly, for the prediction of each target token, given the source sentence $x$ and the target prefix tokens $y_{1:i-1}$, the NMT model predicts the next target token $y_i$ with the probability $P_{\text{NMT}}(y_i|x,y_{1:i-1})$ from the softmax distribution over the vocabulary.

In $k$NN-MT \citep{Khandelwal2020NearestNM}, with a sequence of source tokens and a sequence of target prefix tokens $(s,t_{1:i-1})$ from the in-domain data $\mathcal{D}$, the pre-trained base NMT model outputs the hidden representations $f_{\text{kNN}}(s,t_{1:i-1})$ of the $i$-th target token $t_i$ to construct a datastore. 
The definition of the datastore is as follows:
\begin{align}
\mathcal{(K,V)}&=
\notag
\\
&\left \{ (f_{\text{kNN}}(s,t_{1:i-1}),t_i),
\forall{t_i} \in t \mid (s,t)\in \mathcal{D} \right \}, 
\notag
\end{align}
where $\mathcal{K}$ represents all of the keys, while $\mathcal{V}$ represents all of the corresponding values. 

During inference, given a sequence of the source text which needs to be translated and its generated target prefix tokens, the $k$NN-MT will  first retrieve top-k relevant neighbors from the above datatore based on the Euclidean distances \citep{danielsson1980euclidean} between their contextual hidden representation in the decoder and all the keys based on $f_{\text{kNN}}(x,y_{1:i-1})$. The retrieved set is then converted into the distribution $P_{\text{kNN}}(y_i|x,y_{1,i-1})$ over the vocabulary by,

\begin{align}
P_{\text{kNN}}(y_i|x,y_{1:i-1}) & \propto 
\label{kNNP}
\\
\sum_{(k_i,v_i)}  \mathbbm{1}_{y_j=v_j} &\exp(\frac{-d(k_j,f_{\text{kNN}}(x,y_{1:i-1})}{T}) \notag 
\end{align}
where $j \in [1,k] $, and $k_j,v_j$ are the key and value of the retrieved neighbors respectively. $T$ represents the temperature. 

Finally, the prediction of the next token $y_i$ relies on the interpolation of the predictions from the NMT model and the $k$NN search as follows, 

\begin{align}
P_{\text{comb}}(y_i|x&,y_{1:i-1}) =
\label{finalP}
\\ 
\lambda P_{\text{kNN}}(y_i|&x,y_{1:i-1}) +
(1-\lambda)P_{\text{NMT}}(y_i|x,y_{1:i-1}) \notag
\end{align}
where $\lambda$ is a hyper-parameter for merging the two different distributions. 

\subsection{The \textit{trainable}-$k$NN-MT}
As mentioned in the introduction, we found that $k$NN-MT does not achieve comparable performances with the classic fine-tuning as illustrated in Table \ref{tab:ft_knn_mt}. We hypothesize that the base NMT model trained on out-of-domain data might generate inappropriate contextual representations used by in-domain datastore construction and the $k$NN search algorithm. Conclusively, fine-tuning the base NMT model with the in-domain data would be still necessary when in-domain data is achievable. 

Inspired by both of the advantages and disadvantages of the non-parametric $k$NN-MT, we propose \textit{trainable}-$k$NN-MT which involves the statistics from the $k$NN search as an assistance to scale up gradient into the NMT fine-tuning procedure. It can not only bridge the gap between the NMT model and the $k$NN search at a further step during inference, but enhance vanilla fine-tuning performance as well. Specific details of the \textit{trainable}-$k$NN-MT are displayed in Figure~\ref{framework}. It basically contains two parts: the datastore construction (left) and the NMT model fine-tuning (right). 

At the fine-tuning stage, the datastore is constructed on the in-domain training data with parameters of the NMT model, and it is jointly updated with the NMT fine-tuning. After each certain number of fine-tuning steps, the datastore is re-constructed with updated weights of the NMT model. At each fine-tuning step, given a source sentence x and the ground-truth target prefix tokens $y_{1:i-1}$, the \textit{trainable}-$k$NN-MT retrieves top-$k$ nearest neighbors just as what it does in the $k$NN-MT, and the retrieved set of the $k$NN predictions is converted to into a distribution by Equation \ref{kNNP}, and its statistics assists the NMT model how to do back-propagation with gradient scaling. 

Originally, training a NMT model optimizes the parameters $\theta$ via minimizing the cross entropy loss on the in-domain training dataset $\mathcal{D}$ as follows, 
\begin{equation}
\mathcal{L}=\frac{1}{|\mathcal{D}|}\sum_{(x,y) \in \mathcal{D} } -\log 
P_{\text{NMT}}(y_i|x,y_{1,i-1}{;}\theta). \\
\label{eq:nmtP}
\end{equation}
Instead, for the \textit{trainable}-$k$NN-MT, we generally define a function $g_{\text{$k$NN}}(\cdot)$, which generates the gradient scaling coefficient conditioned on the distribution of the $k$NN predictions. Then, our proposed loss for \textit{trainable}-$k$NN-MT will be as follows,
\begin{equation}
\mathcal{L}=\frac{1}{|\mathcal{D}|}\sum_{(x,y) \in \mathcal{D} } -\log g_{\text{$k$NN}}
P_{\text{NMT}}(y_i|x,y_{1,i-1}{;}\theta), \\
\label{eq:new_loss}
\end{equation}
which is translated into $loss_{\text{\textit{trainable}-$k$NN-MT}}$ in Figure~\ref{framework}. 

In the next subsections, we will explicitly describe three ways to specify the function $g_{\text{$k$NN}}(\cdot)$: (1) a gate mechanism applied on the distribution of $k$NN predictions, (2) the $k$NN ground truth probability and (3) reinforcement learning based on the statistics of $k$NN predictions.

\subsubsection{Gate Mechanism}
In our preliminary observations, when the next target token is a content word with lexical meanings, the distribution of the $k$NN predictions is usually skewed with a remarkable highest probability mass. However, such a phenomenon is not obvious in the translations of grammatical relations, such as function word translations, which results in a flat distribution of the $k$NN outputs. An example is shown in Table~\ref{tab:prob}. The probability of the correct next token "you" is $0.201 + 0.028 = 0.229$, which is not remarkable in the distribution of $k$NN outputs. Motivated by such cases, we hypothesize that the NMT model should learn at a greater extent to improve fluency or styling of the translation. 

\begin{table*}[t]
\normalsize

\centering
{
\begin{tabular}{l | l  l  l  l  l  l  l  l}
\toprule[1pt]	
\multirow{2}{*}{Source} & \multicolumn{8}{p{13.5cm}}{\textit{Ist diese Einstellung aktiv, werden Benachrichtigungen wie zum Beispiel Sperren des Bildschirms oder Änderungen des Profils durch ein passives Meldungsfenster angezeigt.}} \\
\midrule[0.75pt]

\multirow{2}{*}{Reference} & \multicolumn{8}{p{13.5cm}}{\textit{If checked, you will be notified through a passive popup whenever PowerDevil has to notify something, such as screen locking or profile change.}} \\
\midrule[0.75pt]
{Subword} & \textit{T@@}  & \textit{tab}  & \textit{you}   & \textit{noti@@}  & \textit{hin@@}  & \textit{you}  & \textit{promp@@} & \textit{de@@} \\
\midrule[0.75pt]
{Probability} & \textit{\textbf{0.344}} &  \textit{0.236} & \textit{0.201} & \textit{0.096} & \textit{0.054} & \textit{0.028} & \textit{0.022} & \textit{0.019} \\

\bottomrule[1pt]
\end{tabular}
}
\caption{An example of the $k$NN predictions with $k$ = 8 in the German-English IT validation set. The target prefix tokens generated are "If checked,", and the correct next token should be "you". The sub-word candidates in Byte Pair Encoding~\cite{sennrich-etal-2016-neural} are retrieved via $k$NN search with corresponding probabilities. }
\label{tab:prob2}
\end{table*}

  


We utilize the maximum probability in the $k$NN distribution, notated as $\text{M}_{\text{kNN}}$, to design $g_{\text{$k$NN}}(\cdot)$. When the distribution of the $k$NN predictions is flat, and $\text{M}_{\text{kNN}}$ is less than some threshold, $g_{\text{$k$NN}}(\cdot)$ can be specified to be a constant in between 0 and 1. Since $g_{\text{$k$NN}}$ is in $(0,1)$, $-\log g_{\text{$k$NN}}P_{\text{NMT}}(y_i|x,y_{1,i-1}{;}\theta)$ in Equation \ref{eq:new_loss} is larger than the original one in vanilla NMT fine-tuning, which leads to greater gradients for updating the weights of the NMT model. On the contrary, we would keep the original gradient calculations from the NMT fine-tuning. In details, $g_{\text{$k$NN}}(\cdot)$ is defined as follows,
\begin{equation}
g_{\text{$k$NN}} = 
\left\{
\begin{aligned}
c & & {\text{M}_{\text{kNN}} \textless \tau}~ \\
1 & &{\text{M}_{\text{kNN}} \geq \tau}, \\
\end{aligned}
\right. 
\label{eq:tau}
\end{equation}
where $c$ is a constant in between 0 and 1, and the hyper-parameter $\tau$ represents the threshold for $\text{M}_{\text{kNN}}$, which plays the role of a gate in controlling when to push the NMT model to learn translations more greatly. An intuitive way to assign the value of $c$ can be $\lambda$ from the original $k$NN-MT. 

\subsubsection{The $k$NN Ground Truth Probability}
One disadvantage of the gate mechanism is that it would be challenging for us to evaluate how the setting of the threshold $\tau$ would affect the performance of the \textit{trainable}-$k$NN-MT, and investigations via ablation studies are definitely needed when it comes to new domains or new languages. To overcome such a problem, we need to figure out solutions that does not contain any hyper-parameters strongly bound to any specific statistics of $k$NN predictions. 

Inspired by the original cross entropy loss \citep{zhang2018generalized}, one potential solution can be utilizing the probability of the ground truth target token from the distribution of $k$NN predictions. If the probability of the ground truth target word from the distribution of $k$NN predictions is low, it is suitable to enhance the learning of the NMT model, regardless of whether the next target word is a content or function word.

Then, we dynamically set $g_{\text{$k$NN}}$ to be the probability of the ground truth as follows, 
\begin{equation}
g_{\text{$k$NN}} = P_{\text{kNN}}(y_i|x,y_{1:i-1}). 
\label{eq:ground_truth}
\end{equation}

However, this method does not work if $P_{\text{kNN}}(y_i|x,y_{1:i-1})$ is zero, which means the ground truth target word is not retrieved by the $k$NN search for some reason. In such a scenario, we must set a minimum of $g_{\text{$k$NN}}$ to avoid training crash. As we know, the most extreme case for the distribution of $k$NN predictions would be a uniform distribution in which all predictions are equally likely with a probability of $1/k$. Therefore, it is reasonable to set $g_{\text{$k$NN}}$ equal to $1/k$ when the $P_{\text{kNN}}(y_i|x,y_{1:i-1})$ is zero.

\subsubsection{Reinforcement Learning}
The success of non-parametric $k$NN methods in generative models relies on its explicit capability of memorizing the training data which enhances generalizations for domain adaption \citep{khandelwal2020generalization, Khandelwal2020NearestNM} without extra training. It outperforms the base NMT model in terms of quality as well as effectiveness. In our explorations, it may maintain the ability to equip the NMT model to know when to learn more greatly for accurate predictions about unseen contexts of different domains. 

In addition, considering that non-parametric $k$NN search is based on the datastore constructed on the golden labeled training data, it can be regarded as a supervised model of translation prediction. It is essential to try to leverage the gap between the $k$NN search and the predictions from the NMT model by directly optimizing the evaluation measures based the $k$NN search, which is very in line with the spirit of reinforcement learning for structured predictions in generative natural language models \citep{paulus2017deep, sutton2018reinforcement, wu2018study}.

As stated in \citet{wu2018study}, the NMT model can be viewed as an \textit{agent}, which interacts with the \textit{environment} with the previous words $y_{1:i-1}$ and the corresponding contextual representations at each training step. The parameters of the agent define a policy, a conditional probability $P_{\text{NMT}}(y_i|x,y_{1:i-1})$, and the agent will pick an action, that is a candidate word out from the vocabulary, according to the policy. 

Different from the setting of \textit{reward} as BLEU \citep{papineni-etal-2002-bleu} in \citet{wu2018study}, in our \textit{trainable}-$k$NN-MT, the reward for the NMT model is the corresponding probability from the distribution of $k$NN predictions, denoted as $R(\hat{y_{i}},y_{i})$, which is defined by comparing the generate $\hat{y_{i}}$ with the ground-truth
sentence $y_{i}$ in terms of their corresponding probabilities in the distribution of $k$NN predictions. Note that the reward $R(\hat{y_{i}},y_{i})$ is now a token-level reward, a scalar for the generated token $\hat{y_{i}}$, which makes another difference compared with \citet{wu2018study}. 

Therefore, the goal of fine-tuning in such a reinforcement learning framework is to minimize the expected reward as follows,
\begin{align}
&\mathcal{L}= 
\label{eq:rl_loss}
\\
&\frac{1}{|\mathcal{D}|}\sum_{(x,y) \in \mathcal{D} } - R(\hat{y_{i}},y_{i})\log 
P_{\text{NMT}}(y_i|x,y_{1:i-1}), \notag
\end{align}
where $R(\hat{y_{i}},y_{i})$ is defined as, 
\begin{align}
R(\hat{y_{i}},&y_{i})=  
\label{eq:reward}
\\
&\left|P_{\text{kNN}}(\hat{y_i}|x,y_{1:i-1}) - P_{\text{kNN}}(y_i|x,y_{1:i-1})\right|. \notag
\end{align}

When $R(\hat{y_{i}},y_{i})$ is zero, it leads training crash in the current design. It means that either $\hat{y_{i}}$ is correct or both of $\hat{y_i}$ and the ground truth $y_i$ have not been retrieved from the $k$NN search. If it is in the first situation, we will keep the loss calculation from the vanilla NMT fine-tuning for the generated token. Otherwise, $R(\hat{y_{i}},y_{i})$ will be set to $1/k$ with the similar reason stated in the method of the $k$NN ground truth probability. 

\begin{table*}[t]
  \normalsize 
  \centering

{
  \begin{tabular}{ l | c c c c c}
  \toprule[1pt]	

  {Domain} & {IT}& {Medical} & {Law} & {Koran}  
  \\
  \midrule[0.75pt]
     {Train} & {177,792}& {206,804} & {447,696} & {14,979} 
  \\
     {Validation} & {2,000}& {2,000} & {2,000} & {2,000}  
  \\
     {Test} & {2,000}& {2,000} & {2,000} & {2,000}  
  \\
  \midrule[0.75pt]
     {De-En Datastore size} & {3.10M}& {5.70M} & {18.38M} & {0.45M}  
  \\
     {En-De Datastore size} & {3.33M}& {6.13M} & {18.77M} & {0.48M}  
  \\

  \bottomrule[1pt]
  \end{tabular}
 }
  \caption{The datastore size and the number of the parallel sentences in the training, validation, test sets of each domain, separately for German-English (De-En) and English-German (En-De) datasets.}
  \label{tab:data}
\end{table*}

\begin{table*}[h]
\normalsize
  \centering
  \resizebox{\hsize}{!}
{
  \begin{tabular}{l l | c c c c | c}
  \toprule[1pt]
  \textbf{Model} &  \textbf{} & \textbf{IT}& \textbf{Medical} & \textbf{Law} & \textbf{Koran} &\textbf{Avg.}
  \\
  
 \midrule[0.75pt]
 \multicolumn{2}{l}{Base MT} \vline  & 
 38.35 & 40.14 & 45.63 & 16.29  & 35.10
 \\
 \multicolumn{2}{l}{Fine-tuned MT} \vline  &
 47.14 & 57.19 & 61.28 & 22.98 & 47.15
 \\
\midrule[0.75pt]

 \multirow{3}{*}{{Base \textit{trainable}-$k$NN-MT}} 
 &{Gate Mechanism} &
48.63 & 57.81 & 61.42 & 22.77 & 47.66
\\
  & {Ground Truth Prob.} &
48.14 & 57.94 & 62.45 & 22.87 & 47.85
\\

  & {RL} &
47.88 & 57.33 & 62.49 & 23.39 & 47.77
 \\

    \midrule[0.75pt]
 \multirow{3}{*}{ {FT \textit{trainable}-$k$NN-MT}}

 &{Gate Mechanism} &
48.98 & 58.20 & 62.06 & 22.97 & 48.05
 \\
  & {Ground Truth Prob.} &
49.31 & 58.28 & 63.41 & 22.90 & 48.48
 \\

  & {RL} &
49.51 & 58.50 & 63.31 & 23.09 & 48.60
\\
   \bottomrule[1pt]
  \end{tabular}
}
  \caption{Performances of the \textit{trainable}-$k$NN-MT in vanilla fine-tuning, which means there is no integration of $k$NN search during inference on German-English multi-domain test sets. The SacreBLEU scores are averaged along domains for overall comparison. Compared with classic fine-tuning, the overall performance can be improved as much as 1.45 BLEU by the \textit{trainable}-$k$NN-MT.}
  \label{tab:de2en_bleu}
\end{table*}

\begin{table*}[h]
\normalsize
  \centering
  \resizebox{\hsize}{!}
{
  \begin{tabular}{l l| c c c c | c}
  \toprule[1pt]
  \textbf{Model} &  \textbf{} & \textbf{IT}& \textbf{Medical} & \textbf{Law} & \textbf{Koran} &\textbf{Avg.}
  \\
  
 \midrule[0.75pt]
 \multicolumn{2}{l}{Base MT} \vline  & 
 38.35 & 40.14 & 45.63 & 16.29  & 35.10
 \\
 \midrule[0.75pt] 
 \multicolumn{2}{l}
 {$k$NN-MT~\cite{Khandelwal2020NearestNM}} \vline  &  
 46.12 & 54.41 & 61.70 &	21.14  & 45.84
 \\
 \multicolumn{2}{l}{Adaptive $k$NN-MT~\cite{zheng2021adaptive}} \vline  &  
 47.20 & 55.71 & 62.64 &	19.39 & 46.24
 \\
 \multicolumn{2}{l}{CKMT~\cite{wang2022efficient}} \vline  &  
 47.94 & 56.92 & 62.98 &	19.92 & 46.94

\\
 \multicolumn{2}{l}{Robust-$k$NN-MT~\cite{jiang-etal-2022-towards}} \vline  & 
 48.90 & 57.28 & 64.07 & 20.71  &  47.74
 \\
 
  \multicolumn{2}{l}{$k$NN-KD~\cite{yang-etal-2022-nearest}} \vline  &
  — & 56.5 & 61.89 & 24.86 & — 
 \\
    
\midrule[0.75pt]

 \multicolumn{2}{l}{$k$NN-FT-MT} \vline  &
 49.33 & 57.46 & 63.63 & 22.95 & 48.34 
 \\
 
\midrule[0.75pt]

 \multirow{3}{*}{{Base \textit{trainable}-$k$NN-MT}}  & 
 {Gate Mechanism} &
  49.23 & 58.00 & 64.10 & 23.74 & 48.77
 \\

  & {Ground Truth Prob.} &
  49.49 & 58.15 & 64.48 & 23.59  & 48.93 
 \\

  & {RL} &
  49.14 & 57.40 & 64.44 & 23.68 & 48.67
 \\

    \midrule[0.75pt]
 \multirow{3}{*}{ {FT \textit{trainable}-$k$NN-MT}} &
 {Gate Mechanism} &
  {49.96} & {58.34} & {64.67} & {23.81} & {49.20}
 \\

  & {Ground Truth Prob.} &
  49.97 & 58.39 & 64.78 & 23.84  & 49.25
 \\

  & {RL} &
49.84 & 58.60 & 64.99 & 23.78 & 49.30
\\
   \bottomrule[1pt]
  \end{tabular}
}
  \caption{Performances of the \textit{trainable}-$k$NN-MT with the integration of $k$NN search during inference on German-English multi-domain test sets. The FT \textit{trainable}-$k$NN-MT significantly outperforms all of the baseline systems with the $k$NN search algorithm and yields an rough improvement of 1  compared with $k$NN-FT-MT.}
  \label{tab:de2en_knn_bleu}
\end{table*}

\begin{table*}[h]
\normalsize
  \centering
  \resizebox{\hsize}{!}
{
  \begin{tabular}{l l | c c c c | c}
  \toprule[1pt]
  \textbf{Model} &  \textbf{} & \textbf{IT}& \textbf{Medical} & \textbf{Law} & \textbf{Koran} &\textbf{Avg.}
  \\
  
 \midrule[0.75pt]
 \multicolumn{2}{l}{Base MT} \vline & 
 29.74 & 35.56 & 40.85 & 13.97 & 30.03
 \\
 \multicolumn{2}{l}{Fine-tuned MT} \vline  &
 39.70 & 52.5 & 57.16 & 32.45 & 45.45
 \\
\midrule[0.75pt]

 \multirow{2}{*}{{Base \textit{trainable}-$k$NN-MT}} 

  & {Ground Truth Prob.} &
41.17 & 53.38 & 57.75 & 32.45 & 46.19
\\

  & {RL} &
  40.99 & 53.33 & 57.64 & 32.62 & 46.15
 \\

    \midrule[0.75pt]
 \multirow{2}{*}{{FT \textit{trainable}-$k$NN-MT}}
  & {Ground Truth Prob.} &
  41.65 & 54.10 & 58.41 & 32.74 & 46.73
 \\

  & {RL} &
41.53 & 54.36 & 58.34 & 32.70 & 46.73
\\
   \bottomrule[1pt]
  \end{tabular}
}
  \caption{Performances of the \textit{trainable}-$k$NN-MT in vanilla fine-tuning on English-German multi-domain test sets. Compared with classic fine-tuning, the overall performance can be improved as much as 1.28 BLEU by the \textit{trainable}-$k$NN-MT.}
  \label{tab:en2de_bleu}
\end{table*}

\begin{table*}[h]
\normalsize
  \centering
  \resizebox{\hsize}{!}
{
  \begin{tabular}{l l | c c c c | c}
  \toprule[1pt]
  \textbf{Model} &  \textbf{} & \textbf{IT}& \textbf{Medical} & \textbf{Law} & \textbf{Koran} &\textbf{Avg.}
  \\
  
 \midrule[0.75pt]
 \multicolumn{2}{l}{Base MT} \vline & 
 29.74 & 35.56 & 40.85 & 13.97 & 30.03
 \\
 \midrule[0.75pt] 
 \multicolumn{2}{l}
 {$k$NN-MT~\cite{Khandelwal2020NearestNM}} \vline  &  
 36.44 & 49.74 & 55.73 & 25.87 & 41.95
 \\
 \multicolumn{2}{l}{$k$NN-FT-MT} \vline  &
 40.68 & 53.28 & 58.91 & 32.61 & 46.37
 \\
 
\midrule[0.75pt]

 \multirow{2}{*}{{Base \textit{trainable}-$k$NN-MT}} 

  & {Ground Truth Prob.} &
41.22 & 53.53 & 59.03 & 33.53 & 46.83
\\

  & {RL} &
  41.10 & 53.40 & 58.97 & 33.30 & 46.69
 \\

    \midrule[0.75pt]
 \multirow{2}{*}{ {FT \textit{trainable}-$k$NN-MT}}
  & {Ground Truth Prob.} &
  41.73 & 54.50 & 59.21 & 32.99 & 47.11
 \\

  & {RL} &
41.73 & 54.63 & 59.22 & 32.88 & 47.12
\\
   \bottomrule[1pt]
  \end{tabular}
}
  \caption{Performances of the \textit{trainable}-$k$NN-MT with the integration of $k$NN search during inference on English-German multi-domain test sets. The FT \textit{trainable}-$k$NN-MT significantly outperforms all of the baseline systems with the $k$NN search algorithm and yields an improvement of 0.75 BLEU compared with $k$NN-FT-MT.}
  \label{tab:en2de_knn_bleu}
\end{table*}

\section{Experiments}
In this section, we will describe our experimental design and report and discuss experimental results and findings. 

\subsection{Data} 
We conduct experiments of the \textit{trainable}-$k$NN-MT on German-English translation tasks, keeping on the same track as the $k$NN-MT~\cite{Khandelwal2020NearestNM} does, which include the IT, Medical, Law, and Koran domains. We also conduct experiments on English-German translation tasks to evaluation the performance of the \textit{trainable}-$k$NN-MT on from-English translation task. We use the same datasets as German-English translation tasks, but switching the source and target side.

In order to pre-process the data, we perform maximum length filtering with 250 on all of the in-domain training data to ensure data quality. The statistics of the in-domain datasets are shown in Table~\ref{tab:data}. 

\subsection{Experimental Setup}
\paragraph{Pre-trained NMT Models}
In our experimental design, we use the pre-trained winner systems of WMT 2019 German-English and English-German news translation tasks ~\citep{ng-etal-2019-facebook} as the base NMT models for the \textit{trainable}-$k$NN-MT, which are implemented with the Fairseq toolkit~\cite{ott-etal-2019-fairseq} based on the big Transformer architecture \citep{vaswani2017attention}. 

Given the base NMT model pre-trained on the out-of-domain data, we apply the \textit{trainable}-$k$NN-MT algorithm for fine-tuning and compare its performance with the classic vanilla fine-tuning. In addition, when the "pre-trained" NMT model is a fine-tuned model, it is also worth evaluating the performance of \textit{trainable}-$k$NN-MT to see if the NMT model can be  continuously enhanced by the \textit{trainable}-$k$NN-MT algorithm, even though the NMT model has been trained and optimized on the in-domain datasets. For either of the above cases, we conduct experiments of \textit{trainable}-$k$NN-MT with three different ways of gradient scaling and report SacreBLEU results with and without the integration of $k$NN search during inference. 
\paragraph{Model Setting}
The \textit{trainable}-$k$NN-MT is initialized with the pre-trained NMT model and fine-tuned with the Adam algorithm~\cite{DBLP:journals/corr/KingmaB14}. The learning rate is set to 5e-04 or 7e-05 for fine-tuning based on the base NMT model or continuously fine-tuning based on the fine-tuned model respectively. All experiments are run on a single Tesla V-100 GPU card with a batch size of 2048 tokens and a gradient accumulation of 32 batches. The datastore is re-constructed with the updated weights of NMT model after each epoch training, and this procedure repeats until the NMT model converges. 

The hyper-parameters $k$, $\lambda$ and $T$ are tuned on the validation set of each domain, shown in table \ref{tab:de-en-params} and \ref{tab:en-de-params} of appendix \ref{sec:appendix}. We empirically set $\tau$ in Equation~\ref{eq:tau} to $0.6$ for each domain based on the ablation studies. 

\paragraph{Efficiency of $k$NN Search}
In terms of time and storage efficiency, we follow \citet{Khandelwal2020NearestNM} to use FAISS~\cite{Johnson2017BillionscaleSS} index to represent the domain-specific datastore and search for nearest neighbors, with which the keys can be stored in clusters to speed up search and be quantized to 64-bytes for space efficiency, and the index can be constructed offline via a single forward pass over every example in the given in-domain datasets.


\paragraph{Evaluation} We evaluate the performance of \textit{trainable}-$k$NN-MT in the cases with two different pre-trained NMT models as aforementioned and compare them with classic fine-tuning and other competitive models involving non-parametric $k$NN search. The final results are evaluated with SacreBLEU~\cite{post-2018-call} in a case-sensitive detokenized setting~\footnote{We use the exact same evaluation process as the $k$NN-MT does.}.

Among these, (1) \textbf{Base MT} represents the base NMT model, the winner system of WMT 2019 news translation~\cite{ng-etal-2019-facebook} trained with out-of-domain data; 
(2) \textbf{Fine-tuned MT} represents fine-tuning Base MT with the in-domain dataset; 
(3) \textbf{$k$NN-MT} stands for the original $k$NN-MT algorithm ~\cite{Khandelwal2020NearestNM} applied to the Base MT;
(4) \textbf{$k$NN-FT-MT} means the original $k$NN-MT algorithm applied on Fine-tuned MT, where the datastore is constructed with Fine-tuned MT; 
(5) \textbf{Base \textit{trainable}-$k$NN-MT} stands for fine-tuning with our \textit{trainable}-$k$NN-MT from the Base MT;
(6) \textbf{FT \textit{trainable}-$k$NN-MT} means continuously fine-tuning with our \textit{trainable}-$k$NN-MT from Fine-tuned MT. All other competitive systems listed in the tables of results are cited accordingly.

\subsection{Experimental Results}
\paragraph{Results without integration of $k$NN search during inference.}
Our experimental results without the integration of $k$NN search during inference are summarized in Table~\ref{tab:de2en_bleu} and Table \ref{tab:en2de_bleu}. 
For both of German-English and English-German translations, the \textit{trainable}-$k$NN-MT is significantly superior over the classic fine-tuning in both of the Base \textit{trainable}-$k$NN-MT and the FT \textit{trainable}-$k$NN-MT. For German-English translations, the SacreBLEU scores are generally improved by as much as $0.7$ and $1.45$ separately, while for English-German translations, the SacreBLEU scores are improved by $0.74$ and $1.28$ separately. 

Moreover, even though the gate mechanism performs comparably with the $k$NN ground truth probability in both of the the Base \textit{trainable}-$k$NN-MT and the FT \textit{trainable}-$k$NN-MT, it is not an economical method, since ablation studies on the setting of hyper-parameter $tau$ in Equation \ref{eq:tau} are always needed once it comes to a new domain or language. Due to this reason, we performed ablation studies on $tau$ for German-English translations, but we didn't construct such experiments in English-German translations. 

\paragraph{Results with integration of $k$NN search during inference.}
With the integration of $k$NN search during inference, the \textit{trainable}-$k$NN-MT shows its advantages in multiple domains, which is shown in Table \ref{tab:de2en_knn_bleu} and Table \ref{tab:en2de_knn_bleu}. It can be observed that both of Base \textit{trainable}-$k$NN-MT and FT \textit{trainable}-$k$NN-MT significantly outperforms all of the baseline systems involving $k$NN search algorithms. 

Base \textit{trainable}-$k$NN-MT considerably outperforms the original $k$NN-MT and Fine-tuned MT by as much as $3.09$ BLEU and $1.78$ BLEU respectively for German-English translations and $4.88$ BLEU and $1.38$ BLEU respectively for English-German translations. It also outperforms the $k$NN-FT-MT in all of the four domains by an average of $0.61$ BLEU and $0.46$ BLEU for German-English and English-German translations separately. .

Moreover, FT \textit{trainable}-$k$NN-MT generally outperforms the original $k$NN-MT and Fine-tuned MT by as much as $3.46$ BLEU and $2.15$ BLEU respectively for German-English translations and $5.17$ BLEU and $1.67$ BLEU respectively for English-German translations. It also outperforms the $k$NN-FT-MT in all of the four domains by an average of $1$ BLEU and $0.75$ BLEU for German-English and English-German translations separately, which prominently verifies that even if the NMT model has been fine-tuned on the in-domain data, the \textit{trainable}-$k$NN-MT algorithm continues to improve the translation performance consistently. Compared with Base \textit{trainable}-$k$NN-MT, the training of FT \textit{trainable}-$k$NN-MT is more efficient, economical and practical.

\paragraph{Comparisons among methods of gradient scaling.}
Among the three methods of gradient scaling, reinforcement learning framework achieves the best in the FT \textit{trainable}-$k$NN-MT for both of German-English and English-German translations, while it performs comparably with the $k$NN ground truth probability in the Base \textit{trainable}-$k$NN-MT. In the reinforcement learning framework, it is reasonable because the $k$NN search is treated as a supervised reward model. Through preliminary observations, we have found that in retrieving translation candidates, the $k$NN search from the datastore constructed with fine-tuned model weights can achieve higher accuracy than that constructed with base model weights. The better the reward model, the better the effect of reinforcement learning.

We also perform qualitative analysis on the translations from the FT \textit{trainable}-$k$NN-MT compared with $k$NN-FT-MT, since $k$NN-FT-MT was the best model we could get before we proposed the \textit{trainable}-$k$NN-MT. Interestingly, we found that when translating grammatical relations,  FT \textit{trainable}-$k$NN-MT with any of the three gradient scaling methods performs better than $k$NN-FT-MT, which was unexpected. Examples are displayed in Table \ref{tab:case} of Appendix \ref{sec:appendix}. 

\section{Conclusion}
In this paper, we propose the \textit{trainable}-$k$NN-MT to learn translations with the assistance of statistics from the non-parametric $k$NN predictions. It utilizes a gate mechanism, the $k$NN ground truth probability, and reinforcement learning to make full use of the respective advantages and disadvantages of the non-parametric $k$NN search algorithm.  Experimental results show that the \textit{trainable}-$k$NN-MT significantly outperforms the original $k$NN-MT and the classic fine-tuning method, making it a novel fine-tuning method for various domains and translation tasks. 




\bibliography{anthology,custom}
\bibliographystyle{acl_natbib}

\appendix
\section{Appendix}
\label{sec:appendix}

\begin{table*}[t]
\normalsize

\centering
{
\begin{tabular}{l | l | l p{1.1cm}  p{1.1cm}  p{1.1cm} p{1.1cm}  p{1.1cm}  p{1.1cm}  p{1.1cm}  p{1.1cm}}
\toprule[1pt]	
\multirow{2}{*}{Source} & \multicolumn{9}{p{13cm}}{\textit{Sie können Writer-Textrahmen so miteinander verketten, dass ihr Inhalt automatisch von einem Rahmen in den nächsten fließt.}} \\
\midrule[0.75pt]

\multirow{2}{*}{Reference} & \multicolumn{9}{p{13cm}}{\textit{You can link Writer text frames so that their contents automatically flow from one frame to another.}} \\
\midrule[0.75pt]
{Target Prefix} & \multicolumn{9}{l}{\textit{You can}}   \\
\midrule[0.75pt]

\multirow{2}{*}{$k$NN-MT} 
& {Sub.} & \textit{join} & \textit{link} & \textit{chain} & \textit{link} & \textit{use} & \textit{link} & \textit{connect} & \textit{comb@@} \\
& {Prob.} & \textit{0.379} & \textit{0.242} & \textit{0.103} & \textit{0.093} & \textit{0.068} & \textit{0.041} & \textit{0.037} & \textit{0.032} \\
\midrule[0.75pt]

\multirow{2}{*}{$k$NN-FT-MT}
& {Sub.} & \textit{link} & \textit{chain} & \textit{nest} & \textit{nest} & \textit{link} & \textit{link} & \textit{link} & \textit{link} \\
& {Prob.}  & \textit{0.698} & \textit{0.243} & \textit{0.029} & \textit{0.012} & \textit{0.008} & \textit{0.003} & \textit{0.002} & \textit{0.001}  \\

  \midrule[0.75pt]
  \midrule[0.75pt]

\multirow{2}{*}{Source} & \multicolumn{9}{p{13cm}}{\textit{Die Quell- und Zielansicht ist der Hauptarbeitsbereich von \& kompare;. Hier werden der Inhalt und die hervorgehobenen Abweichungen der aktuell ausgewählten Quell- und Zieldatei mit den Zeilennummern angezeigt.
}} \\

\midrule[0.75pt]

\multirow{2}{*}{Reference} & \multicolumn{9}{p{13cm}}{\textit{The source and destination view is the main workspace of \& kompare;. The contents and highlighted differences of the currently selected source and destination file are displayed here with line numbers.}} \\

\midrule[0.75pt]

{Target Prefix} & \multicolumn{9}{l}{\textit{The source and destination}}   \\
\midrule[0.75pt]

\multirow{2}{*}{$k$NN-MT} 
& {Sub.} & \textit{p@@}  & \textit{ann@@}  & \textit{brow@@}   & \textit{view}  & \textit{view}  & \textit{view}  & \textit{view} & \textit{view} \\
& {Prob.} & \textit{0.679} & \textit{0.110} & \textit{0.049} & \textit{0.039} & \textit{0.035} & \textit{0.034} & \textit{0.025} & \textit{0.025} \\
\midrule[0.75pt]

\multirow{2}{*}{$k$NN-FT-MT}
& {Sub.} & \textit{view}  & \textit{View}  & \textit{view}   & \textit{View}  & \textit{View}  & \textit{View}  & \textit{view} & \textit{view} \\
& {Prob.} & \textit{0.361} & \textit{0.193} & \textit{0.159} & \textit{0.073} & \textit{0.073} & \textit{0.048} & \textit{0.047} & \textit{0.043}  \\

\bottomrule[1pt]
\end{tabular}
}
\caption{Examples of the $k$NN predictions in German-English IT domain translation task, where $k$NN-FT-MT means applying $k$NN-MT algorithm on the fine-tuned NMT model, Sub. and Prob. represent $k$NN retrieved tokens and corresponding probabilities in the $k$NN distribution and Target Prefix stands for prefix generated target tokens. The $k$NN search algorithm secures more accurate prediction candidates when the NMT model has been fine-tuned with the IT training data. }
\label{tab:prob}
\end{table*}

\begin{table*}[t]
  \normalsize 
  \centering

{
  \begin{tabular}{ l | l | c c c c c}
  \toprule[1pt]	

  {Model} & & {IT}& {Medical} & {Law} & {Koran}  
  \\
  \midrule[0.75pt]
   \multirow{3}{*}{$k$NN-MT \& Base \textit{trainable}-$k$NN-MT} &   {$k$} & {8}& {16} & {16} & {8}  
  \\
    & {$\lambda$} & {0.6}& {0.8} & {0.8} & {0.6}  
  \\
    & {$T$} & {5}& {5} & {5} & {100}  
  \\
  \midrule[0.75pt]
   \multirow{3}{*}{$k$NN-FT-MT \& FT \textit{trainable}-$k$NN-MT} &   {$k$} & {8}& {16} & {8} & {8}   
  \\
    & {$\lambda$} & {0.4}& {0.4} & {0.6} & {0.4}  
  \\
    & {$T$} & {10}& {10} & {10} & {100}  
  \\

  \bottomrule[1pt]
  \end{tabular}
 }
  \caption{The hyper-parameters used in the $k$NN based models of German-English translations.}
  \label{tab:de-en-params}
\end{table*}

\begin{table*}[t]
  \normalsize 
  \centering

{
  \begin{tabular}{ l | l | c c c c c}
  \toprule[1pt]	

  {Model} & & {IT}& {Medical} & {Law} & {Koran}  
  \\
  \midrule[0.75pt]
   \multirow{3}{*}{$k$NN-MT \& Base \textit{trainable}-$k$NN-MT} &   {$k$} & {8}& {8} & {16} & {16}  
  \\
    & {$\lambda$} & {0.6}& {0.8} & {0.8} & {0.8}  
  \\
    & {$T$} & {10}& {10} & {5} & {10}  
  \\
  \midrule[0.75pt]
   \multirow{3}{*}{$k$NN-FT-MT \& FT \textit{trainable}-$k$NN-MT} &   {$k$} & {4}& {4} & {8} & {16}  
  \\
    & {$\lambda$} & {0.4}& {0.4} & {0.4} & {0.2}  
  \\
    & {$T$} & {10}& {100} & {5} & {5}  
  \\

  \bottomrule[1pt]
  \end{tabular}
 }
  \caption{The hyper-parameters used in the $k$NN based models of English-German translations.}
  \label{tab:en-de-params}
\end{table*}

\begin{table*}[t]
\normalsize

  \centering
{
  \begin{tabular}{l | p{11cm}}
 \toprule[1pt]	

   \multirow{2}{*}{Source} &  \textit{Sollte Seine Peinigung über euch nachts oder am Tage hereinbrechen, was wollen denn die schwer Verfehlenden davon beschleunigen?}
  \\
  \midrule[0.75pt]
  {Reference} & \textit{If His chastisement comes upon you by night or day, what part of it will the sinners seek to hasten?}
 \\
 \midrule[0.75pt]
 {$k$NN-FT-MT} & \textit{If His punishment befalls you at night or in the day, what would the sinners do to despatch it?} 
 \\
\midrule[0.75pt]
  {Gate Mechanism (Ours)} & \textit{If His punishment \textbf{comes upon} you \textbf{by night or by day}, how will the sinners hasten it?} 
\\

\midrule[0.75pt]
  {Ground Truth Prob. (Ours)} & \textit{If His punishment \textbf{comes upon} you at night or in the day, how will the sinners hasten it?} 
\\

\midrule[0.75pt]
  {RL (Ours)} & \textit{If His punishment \textbf{comes upon} you at night or in the day, \textbf{what} will the sinners do to hasten it?} 
\\

  \midrule[0.75pt]
  \midrule[0.75pt]

  \multirow{2}{*}{Source} & \textit{Und sie sagen: "Wir glauben daran." Aber wie könnten sie (den Glauben)  von einem fernen Ort aus erlangen,}
  \\
  \midrule[0.75pt]
  {Reference} & \textit{and they say, `We believe in it'; but how can they reach from a place far away,}
 \\
 \midrule[0.75pt]
 {$k$NN-FT-MT} & \textit{They say: "We believe in it;" but how could they reach it from a place of no return?}
 \\
\midrule[0.75pt]
 {Gate Mechanism (Ours)} & \textit{\textbf{And they say}, "We believe in it"; so how can they reach it from a place \textbf{far away?}}
\\

\midrule[0.75pt]
 {Ground Truth Prob. (Ours)} & \textit{\textbf{And they say}, "We believe in it"; so how can they reach it from a place of no return?}
\\

\midrule[0.75pt]
 {RL (Ours)} & \textit{\textbf{And they say}, "We believe in it"; so how can they reach it from a place of no return?}
\\

\midrule[0.75pt]
\midrule[0.75pt]
 {Source} &  \textit{Und haltet sie an; denn sie sollen befragt werden.}
  \\
  \midrule[0.75pt]
  {Reference} & \textit{And detain them, for they will be questioned.}
 \\
 
 \midrule[0.75pt]
  {$k$NN-FT-MT} & \textit{Surely they are to be interrogated.} 
 \\
\midrule[0.75pt]
  {Gate Mechanism (Ours)} & \textit{And test them, and they \textbf{will be} questioned.} 
\\
\midrule[0.75pt]
  {Ground Truth Prob. (Ours)} & \textit{Persevere with them, and they \textbf{will be} questioned.} 
\\
\midrule[0.75pt]
  {RL (Ours)} & \textit{Persevere with them, and they \textbf{will be} questioned.} 
\\


  \bottomrule[1pt]
  \end{tabular}
}
  \caption{Examples of translations from the \textit{trainable}-$k$NN-MT with three ways to gradient scaling compared with \textbf{$k$NN-FT-MT} in the Koran domain.  Boldfaced words indicate their differences. The \textbf{FT \textit{trainable}-$k$NN-MT} can translate correctly in the case of grammatical relation translations, while the \textbf{$k$NN-FT-MT} can't.}
  
  \label{tab:case}
\end{table*}

\end{document}